\documentclass{article}
\usepackage{tipa}
\usepackage{spconf,amsmath,graphicx,hyperref}
\usepackage{multirow}
\usepackage{graphicx}
\usepackage{microtype}
\usepackage{inconsolata}
\usepackage[utf8]{inputenc}
\usepackage[T1]{fontenc}
\usepackage{times}
\usepackage{latexsym}
\usepackage{booktabs}
\usepackage{float}
\usepackage{xcolor}
\usepackage{amsmath}   
\usepackage{amssymb}
\usepackage{etoolbox}
\apptocmd{\thebibliography}{\setlength{\itemsep}{1.2pt}\setlength{\parskip}{0.4pt}}{}{}

\title{Towards Fair ASR for Second Language Speakers Using Fairness Prompted Finetuning}

\name{
\begin{tabular}{c}
Monorama Swain$^{1}$, Bubai Maji$^{2}$, Jagabandhu Mishra$^{3}$, Markus Schedl$^{1}$, \\ Anders Søgaard$^{4}$,  Jesper Rindom Jensen$^{5}$ 
\end{tabular} }

\address{
$^{1}$Johannes Kepler University Linz, Austria,
$^{2}$IIT Kharagpur, India,
$^{3}$University of Eastern Finland, \\
$^{4}$University of Copenhagen, Denmark, 
$^{5}$Aalborg University, Denmark  
}

\begin{document}
%
\maketitle
\begin{abstract}
In this work, we address the challenge of building fair English ASR systems for second-language speakers. Our analysis of widely used ASR models, Whisper and Seamless-M4T, reveals large fluctuations in word error rate (WER) across $26$ accent groups, indicating significant fairness gaps. To mitigate this, we propose fairness-prompted finetuning with lightweight adapters, incorporating Spectral Decoupling (SD), Group Distributionally Robust Optimization (Group-DRO), and Invariant Risk Minimization (IRM). Our proposed fusion of traditional empirical risk minimization (ERM) with cross-entropy and fairness-driven objectives (SD, Group DRO, and IRM) enhances fairness across accent groups while maintaining overall recognition accuracy. In terms of macro-averaged word error rate, our approach achieves a relative improvement of $58.7\%$ and $58.5\%$ over the large pretrained Whisper and Seamless-M4T, and $9.7\%$ and $7.8\%$ over them, finetuning with standard empirical risk minimization with cross-entropy loss.

\end{abstract}
\section{Introduction}
\label{sec:intro}
Speech technologies can be a driver of equality, making information accessible to social groups with limited access: dyslexics, non-literates, congenitally blind, children, the elderly, and language learners, if they work well across all social groups, including non-native speakers \cite{tan2007acoustic}. Performance drops for non-native speakers can often be traced back to the influence of the speaker’s mother tongue and its effect on pronunciation \cite{livescu2000lexical}, \cite{oh2007acoustic}. Accents may exhibit a lot of variation: In Indian English, for example, the /z/ sound is often replaced by /s/, especially in the speech of those whose native languages do not distinguish between the two. Moreover, \textit{very} might be pronounced as [\textipa{"Ve:Ri}], using the retroflex /ó/ instead of the alveolar /r/ \cite{gargesh2008indian}.



\begin{figure}[!t]
\centering
\includegraphics[width=\columnwidth]{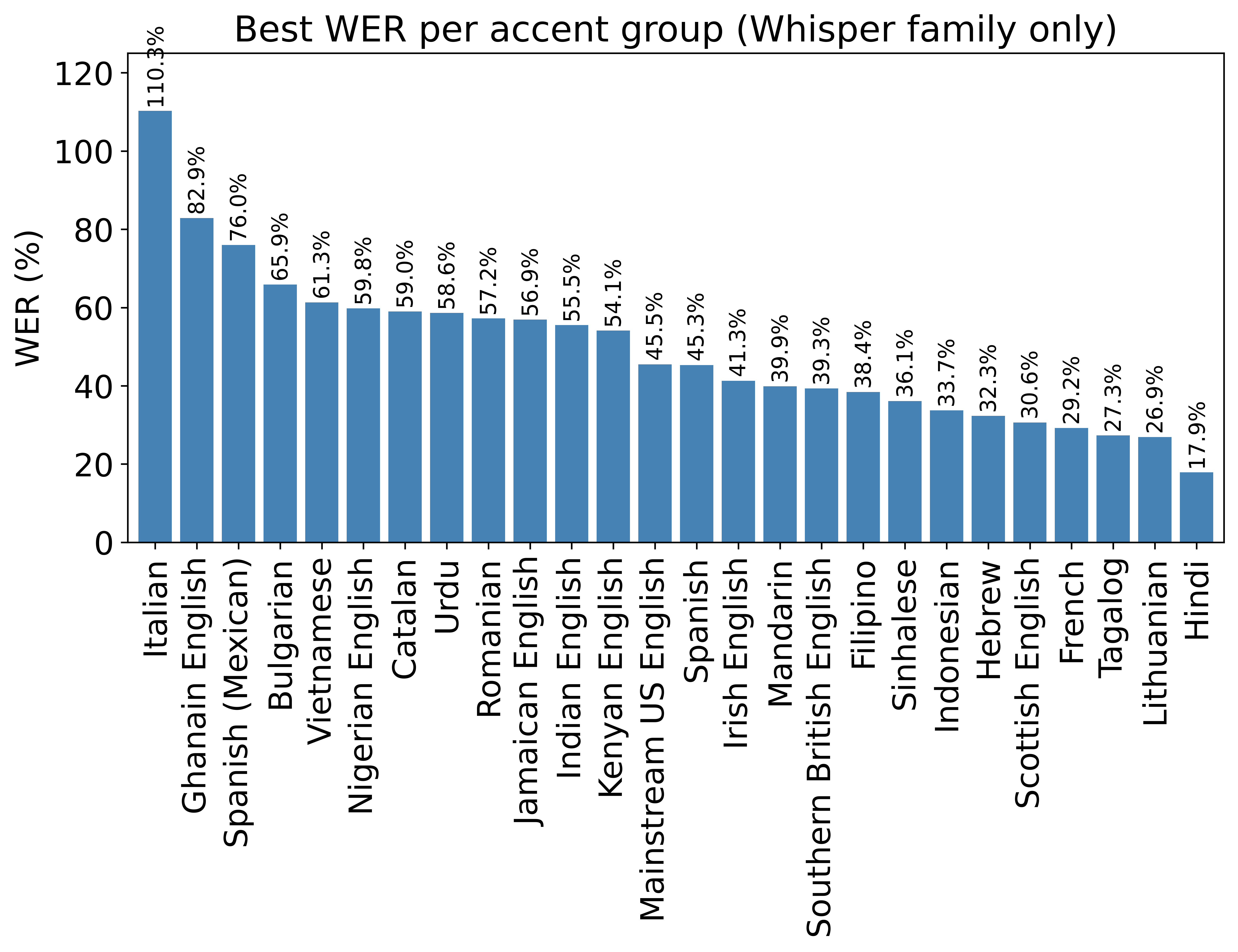}
\vspace{-0.6 cm}
  \caption{Best WER (\%) per accent group for Whisper models before fine-tuning. For each group, the WER shown corresponds to the lowest WER obtained among Whisper variants (i.e., tiny, base, small, medium, large)}
  \vspace{-0.4 cm}
  \label{fig:figure1}
\end{figure}

Automatic speech recognition (ASR) systems have achieved remarkable performance in high-resource languages such as English. Yet, several studies have shown that these systems exhibit unequal performance across demographic and accent groups \cite{dheram2022toward, koenecke2020racial, harris2024modeling, jahan2025unveiling, joshi25_interspeech}, a disparity further illustrated in Figure~\ref{fig:figure1}. Such fairness gaps highlight the pressing need for systematic methodologies to evaluate and mitigate bias, aligning with the call in \cite{joshi25_interspeech} that “Every voice matters.” A key reason for these disparities is the over-representation of mainstream accents in training and evaluation data \cite{dichristofano2022global}, leading to substantial performance gaps for underrepresented accents.

In this paper, we focus on improving the fairness and robustness of ASR across various English accents, with particular attention to \emph{second-language (L2)} speakers. There has been some attempts already, however, to study fairness in ASR. For example, in \cite{liu2022model}, a mixed-effects poisson regression has been introduced as a statistical approach to measure and interpret word error rate (WER) differences among subgroups. This method has been shown to address challenges such as handling unobserved heterogeneity across speakers and identifying the sources of WER gaps between subgroups. Similarly, \cite{sari2021counterfactually} introduced counterfactual training methods for Connectionist Temporal Classification-based ASR, where demographic information such as gender, age, education are counterfactually modified while the linguistic content remains fixed. Their method helps to reduce the variance in character error rate across demographic groups. 

\begin{figure}[!t]
  \centering
  \includegraphics[trim= 10mm 92mm 38mm 88mm, clip,width=1.0\linewidth]{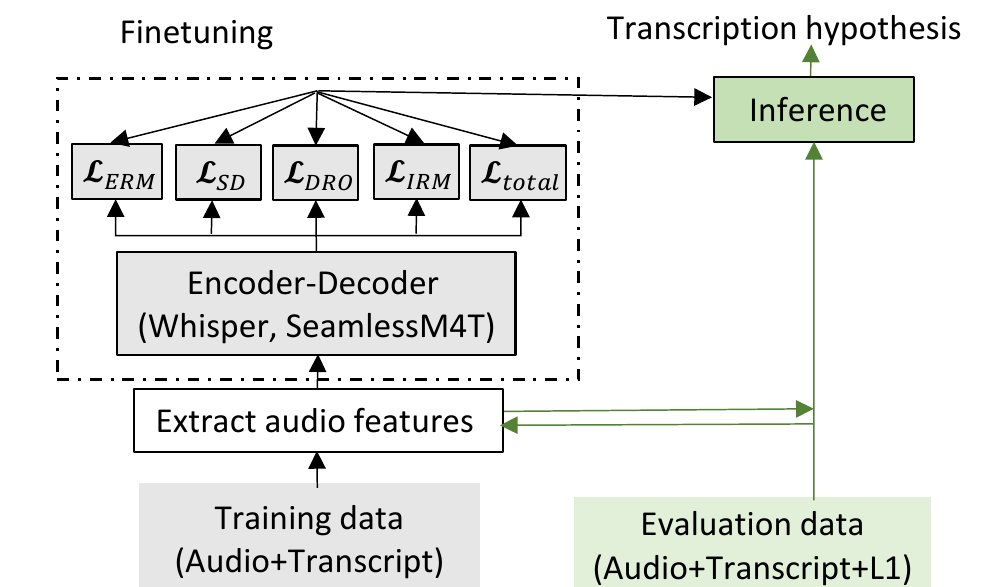}
  \caption{The fairness-prompting speech recognition system. }
  \label{fig:figures2}
\end{figure}

Recent work, such as \cite{lee2025ml}, has benchmarked ASR performance across languages and accents, highlighting performance disparities of foundational ASR models across accent groups. Similarly, \cite{dheram2022toward} leveraged one-hot accent embeddings to improve adaptation to diverse speakers. However, prior studies have primarily focused on modeling accent variation, and to the best of our knowledge, none have explored \emph{fairness prompting} as a means to fine-tune foundational models. Our work addresses this gap by explicitly targeting \emph{L2 English} varieties and incorporating \emph{fairness disparities} into the learning objective, thereby enabling \emph{fair finetuning}. Our main contributions are:
\begin{itemize}
    \item We conducted systematic fairness analyzes targeting L2 English speakers, highlighting disparities that remain hidden when treating all accents uniformly.
    \item We evaluated two families of off-the-shelf speech models (Whisper and SeamlessM4T) under standard vanilla fine-tuning (i.e., empirical risk minimization (ERM)) as the baseline with the fairness-promoting algorithms: Group-DRO; spectral decoupling (SD); and invariant risk minimization (IRM).
    \item Finally, we propose a novel fusion of ERM, SD, Group-DRO, and IRM (see Figure~\ref{fig:figures2}) to evaluate its effectiveness in reducing performance gaps across $26$ L2 English accents. In addition, we also analyze how model size and linguistic/data properties influence the effectiveness of fairness-promoting fine-tuning. 
    
\end{itemize}



 
\section{Methodology}

\subsection{Problem Statement}
Our goal is to minimize the disparities in ASR performance across $26$ groups. Each group contains English utterances shaped by a distinct L2 corresponding to one native language. Thus, each sample consists of the information ($x$, $y$, $g$),
where $x \in \mathbb{R}^T$ is the speech sample, where $T$ is the length of the input speech signal,
$y = (y_1, \dots, y_M)$ is the English transcription sequence, and 
$g \in \{1,\dots,26\}$ denotes the accent group.

\vspace{-0.3cm}
\subsection{ASR Model}
We adopt large-scale pretrained sequence-to-sequence ASR models, i.e.,  Whisper family and SeamlessM4T: 

\textbf{Whisper \cite{radford2023robust}:} A family set of five ASR models (tiny, base, small, medium, and large)\footnotemark, which are trained in the same way, but of different sizes, developed by OpenAI. \footnotetext{All Whisper model variants are available at \url{https://huggingface.co/models?search=openai/whisper}.} The Whisper models are
trained on 680,000 hours of data sourced from the web in 97 languages. During training, Whisper uses data augmentation techniques that transform audio spectrograms by applying methods like time warping, frequency masking, and time masking. These techniques enhance the model’s ability to generalize across various acoustic environments.
Whisper is built as a traditional encoder-decoder system and is trained in a weakly supervised manner on unreleased multilingual audio data. To handle long audio, the system divides the audio into 30-second chunks for processing.

\textbf{SeamlessM4T \cite{barrault2023seamlessm4t}:} is a model developed by MetaAI that supports speech-to-speech translation, speech-to-text translation,
text-to-speech translation, text-to-text translation, and ASR for up to 100 languages. The model
has two versions \footnotemark: Medium with 1.2 billion parameters and a size of 6.4 GB, and Large with 2.3 billion parameters and a size of 10.7 GB. \footnotetext{Both Seamless models are available at \url{https://huggingface.co/models?search=seamless}.}


\vspace{-0.3cm}
\subsection{Finetuning with ERM}

ERM \cite{donini2018empirical}: The ERM baseline minimizes the empirical mean cross-entropy over the pooled training set, which helps to maximize the likelihood of the reference transcription.
Accordingly, the ERM objective is formulated as:
\vspace{-0.25cm}
\begin{equation}
\mathcal{L}_{\text{ERM}} = \frac{1}{N} \sum_{n=1}^N \mathcal{L}^{(n)}_{\text{ASR}}
\end{equation}
where $N$ is the total number of training utterances and $\mathcal{L}^{(n)}_{\text{ASR}}$ is the ASR loss (cross-entropy) for the $n$-th utterance. This standard training objective, which optimizes average loss, tends to favor majority groups and can result in higher WER for underrepresented accents.

\vspace{-0.3cm}
\subsection{Fairness-Promoting Finetuning}
To adapt the considered models fairly across groups, we fine-tune with a multi-objective fairness loss. This allows adaptation to non-native English accents.

\textbf{SD \cite{pezeshki2021gradient}:}
SD is a regularization method that penalizes the squared logit magnitude.  It helps to reduce overconfident predictions and spurious correlations to improve generalization across accents. The SD loss is computed as:
\begin{equation}
\mathcal{L}_{\text{SD}} = \mathcal{L}_{\text{ERM}} + \lambda \, \lVert o \rVert_2^2 
\end{equation}
where $o$ are the model logits before the Softmax layer and $\lambda$ is a regularization coefficient. 

\textbf{Group-DRO \cite{Sagawa*2020Distributionally}:}
 Standard ERM minimizes average risk, which can improve majority accents while ignoring minority groups. Group-DRO shifts optimization to the \emph{worst-performing group}, explicitly improving fairness. This can be mathematically expressed as:
\begin{equation}
\mathcal{L}_{\text{DRO}} = \max_{g \in G} \mathcal{L}^{(g)}_{\text{ASR}}
\end{equation}
where $G$ is the set of accent groups and $\mathcal{L}^{(g)}_{\text{ASR}}$ is the average ASR loss within group $g$.

\textbf{IRM \cite{rosenfeld2021the}:}
IRM encourages the model to learn consistent predictors across environments (e.g., channel conditions or demographic metadata) to reduce spurious correlations. The IRM objective is defined as:
\begin{equation}
\mathcal{L}_{\text{IRM}} = \sum_{e \in E} \lVert \nabla_{w} \mathcal{L}^{(e)}_{\text{ASR}} (w \cdot \Phi) \rVert^2
\end{equation}
where $E$ denotes the set of environments, $L^{(e)}_{\text{ASR}}$ the ASR loss in environment $e$, $\Phi$ the learned features, $w$ is a scalar classifier.

\textbf{Fusion:}
The fusion objective reduces performance gaps between accent groups, yielding a fairer ASR system for English as L2. The overall objective is a weighted combination of all terms and is computed as:
\begin{equation}
\mathcal{L}_{\text{total}}
= \lambda_{\mathrm{e}} \mathcal{L}_{\mathrm{ERM}}
+ \lambda_{\mathrm{s}} \mathcal{L}_{\mathrm{SD}}
+ \lambda_{\mathrm{d}} \mathcal{L}_{\mathrm{DRO}}
+ \lambda_{\mathrm{i}} \mathcal{L}_{\mathrm{IRM}}
\end{equation}
where $\lambda_{\mathrm{e}}, \lambda_{\mathrm{s}}, \lambda_{\mathrm{d}}, \lambda_{\mathrm{i}}$ 
are scalar weights that balance the contributions of each loss.


\vspace{-0.3cm}
\section{Experimental Setup and Results}


\subsection{Dataset}
We used the Edinburgh International Accents of English Corpus (EdAcc) \cite{sanabria2023edinburgh} – a new ASR dataset composed of 40 hours of English dyadic conversations between speakers with a diverse set of accents. EdAcc contains more than 40 self-reported English accents from speakers of 51 different first languages. It includes 26 distinct first- and second-language varieties of English, along with a linguistic background profile of each speaker, detailing how long they have spoken English and where they have lived. The conversations range from 20-60 minutes in duration. In our experiments, we used standard data splits as described in the dataset description~\cite{sanabria2023edinburgh}.

\vspace{-0.4cm}
\subsection{Experimental Setup}
For a fair comparison, we evaluated all four fine-tuning strategies, including our baseline ERM approach. We set $\lambda_\mathrm{e}=\lambda_\mathrm{d}=1$ to retain empirical risk and Group-DRO as the primary optimization, while $\lambda_\mathrm{s}=0.06$ and $\lambda_\mathrm{i}=0.01$ were selected through a greedy search in the range $0.01$–$1$. The learning rate was fixed at $4 \times 10^{-5}$. These hyperparameters were tuned on validation performance to balance fairness improvements with ASR accuracy. The ERM baseline adapter was trained on top of the pre-trained Whisper model, enabling adaptation to the EdAcc dataset.

\vspace{-0.3cm}
\subsection{Fairness-Oriented Evaluation Metrics}
To quantify fairness across accent groups, we report (i) the Word Error Rate for each accent group (\text{WER}($g$)), (ii) the Macro-average of \text{WER}($g$) \cite{das2021best}, showing the overall ASR accuracy, averaged fairly across groups (see Equation~\ref{eq:equation6}), and (iii) the Min-Max gap of \text{WER}($g$) \cite{veliche2024towards}, to evaluate the disparity between the worst and best performing groups (see Equation~\ref{eq:equation7}).

\vspace{-0.5cm}
\begin{align}
\text{Macro-average} &= \frac{1}{|G|} \sum_{g \in G} \text{WER}(g)
\label{eq:equation6}\\
\text{Min-Max Gap} &= \max_{g \in G} \, \text{WER}(g)-\min_{g \in G} \, \text{WER}(g)
\label{eq:equation7}
\end{align}
where $G$ is the total number of accent groups (=26).

\begin{table}[!t]
\centering
\scriptsize
\setlength{\tabcolsep}{3pt} 
\renewcommand{\arraystretch}{1.1} 
\caption{Comparison of (macro-average) / (min–max difference) of WER across Whisper and Seamless models under different fine-tuning strategies.}
\begin{tabular}{lcccccc}
\toprule
\textbf{Model} & \textbf{w/o FT} & \textbf{ERM} & \textbf{DRO} & \textbf{SD} & \textbf{IRM} & \textbf{Fusion} \\
\midrule
\multicolumn{7}{l}{\textbf{Whisper}} \\
\midrule
\textbf{Large}  & 58.3 / 114.0 & 26.7 / \textbf{30.1} & 33.3 / 37.6 & 35.7 / 31.5 & 38.3 / 53.2 & \textbf{24.1} / 30.8 \\
\textbf{Medium} & 61.1 / 92.6  & 31.8 / 47.6 & 38.2 / 48.3 & 40.2 / 55.4 & 42.6 / 59.9 & \textbf{28.2} / 35.3 \\
\textbf{Small}  & 67.8 / 120.5 & 32.9 / \textbf{39.4} & 38.3 / 41.9 & 41.4 / 44.3 & 43.8 / 53.2 & \textbf{30.3} / 45.1 \\
\textbf{Base}   & 65.5 / 103.6 & 38.7 / 47.9 & 44.3 / \textbf{43.0} & 48.9 / 57.1 & 41.4 / 59.2 & \textbf{33.7} / 59.8 \\
\textbf{Tiny}   & 96.0 / 124.0 & 48.9 / \textbf{54.1} & 51.4 / 56.9 & 57.8 / 64.0 & 55.7 / 85.1 & \textbf{41.0} / 57.8 \\
\bottomrule
\multicolumn{7}{l}{\textbf{Seamless}} \\
\midrule
\textbf{Large}  & 65.3 / 52.7 & 29.4 / 43.3 & 28.1 / 36.0 & 26.3 / \textbf{28.5} & 32.8 / 48.6 & \textbf{27.1} / 37.6 \\
\textbf{Medium} & 67.2 / 42.5 & 40.5 / 50.8 & \textbf{26.8} / 38.2 & 28.3 / 34.2 & 36.3 / 48.7 & 29.0 / \textbf{29.0} \\
\bottomrule
\end{tabular}

\label{tab:table1}
\end{table}

\begin{figure*}[!t]
\centering
\begin{minipage}{0.8\columnwidth}
  \centering
  \includegraphics[width=\linewidth]{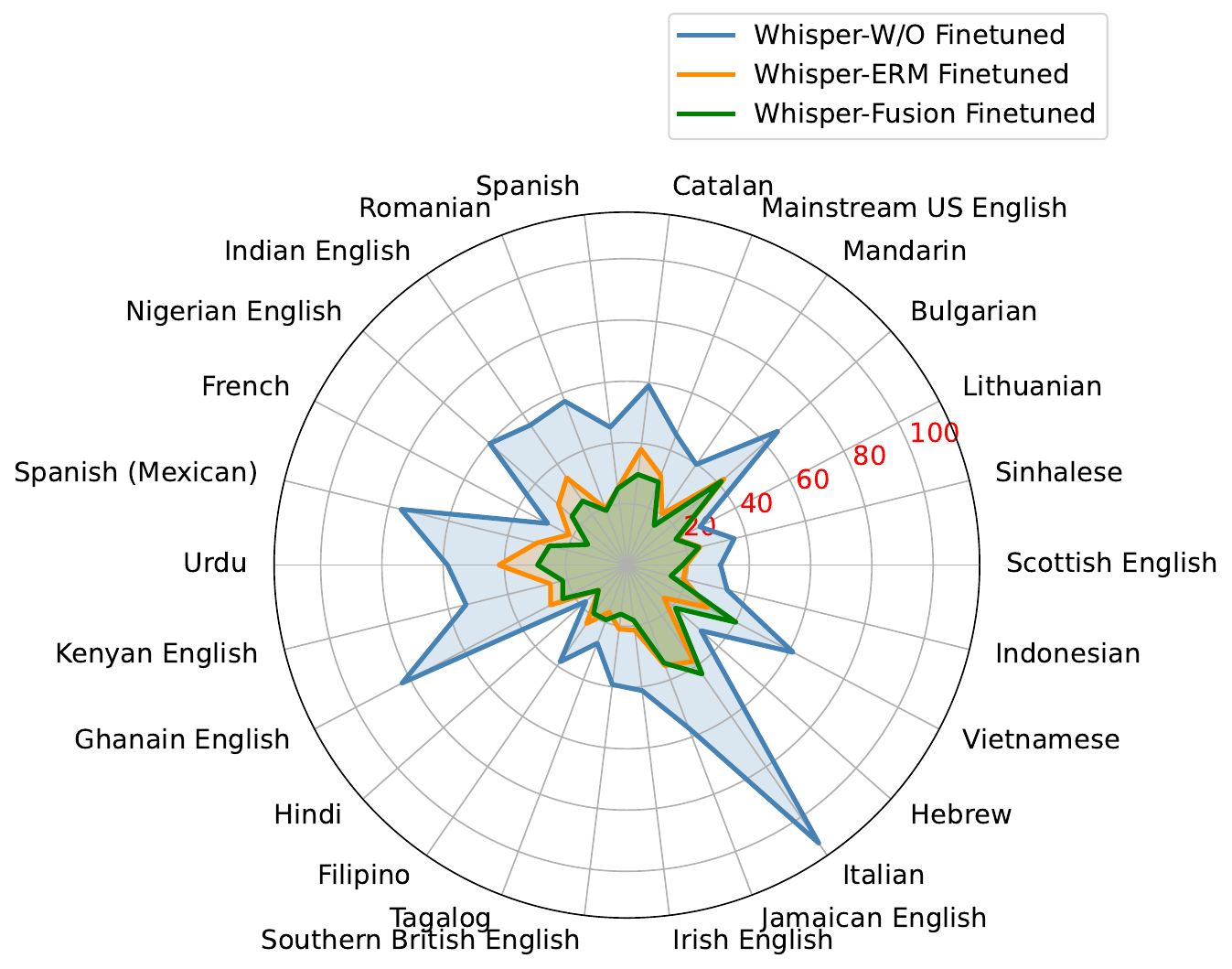}
\end{minipage}
\hspace{0.16\columnwidth}
\begin{minipage}{0.80\columnwidth}
  \centering
  \includegraphics[width=\linewidth]{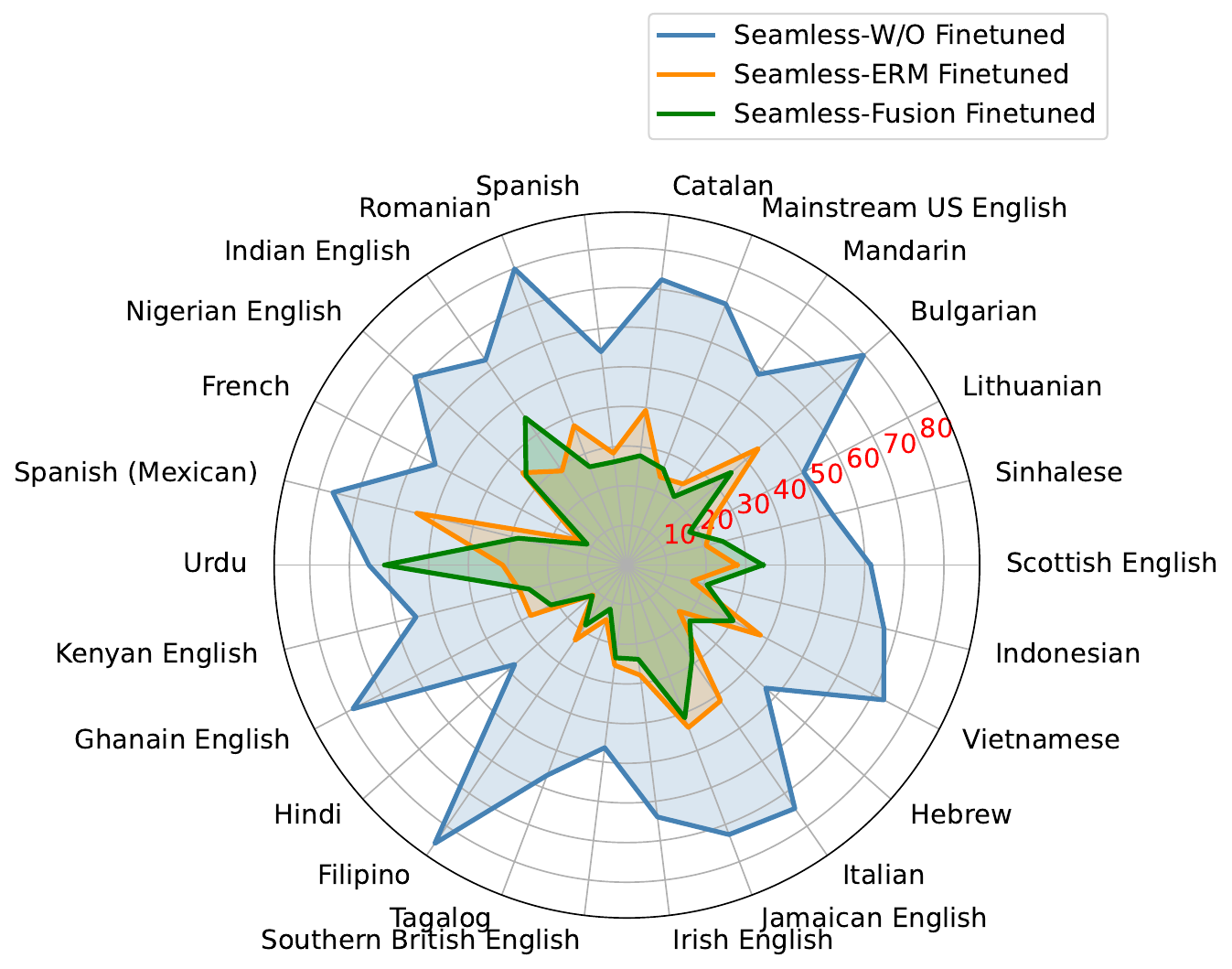}
\end{minipage}
\caption{WER (\%) across accent groups for Whisper (left) and Seamless (right) models under different fine-tuning strategies. Without finetuning (blue) show high WER and disparities. ERM (orange) reduces errors but retains variability, while Fusion (green) yields the most balanced profiles, lowering error rates for challenging accents and improving fairness across groups.}
\label{fig:radar_plots}
\end{figure*}

\vspace{-0.3cm}
\subsection{Results and Discussion}

Table~\ref{tab:table1} reports the macro-average WER and min–max disparities for the Whisper and Seamless families under different fine-tuning strategies. Without fine-tuning, both families perform poorly on accented English. Whisper yields particularly high WERs ($58$–$96\%$) and large disparities ($92$–$124\%$), while Seamless shows smaller disparities ($42$–$53\%$) at baseline, reflecting stronger multilingual pretraining. Fine-tuning consistently improves performance. ERM reduces WER by nearly $50\%$ relative to the baseline—for instance, Whisper-large drops from $58.3\%$ WER to $26.7\%$. Fairness-oriented methods such as DRO and SD further reduce disparities, with Seamless-large under SD achieving the smallest gap ($28.5\%$).

Our proposed fusion objective outperforms the individual fairness-oriented objectives. Across all Whisper architectures, fusion lowers error rates across accents, achieving the best performance with Whisper-large at $24.1\%$ WER. In Seamless, fusion produces a more balanced profile; for example, Seamless-Medium fusion achieves $29\%$ WER and a $29\%$ min–max gap, compared to Seamless-large with $27.1\%$ WER and a larger $37.6\%$ gap. Thus, Whisper provides stronger performance in terms of macro-average WER, whereas Seamless maintains smaller min–max disparities across accents.

Figures~\ref{fig:radar_plots} illustrate per-accent WER for Whisper and Seamless. Without fine-tuning, both models show large spikes for under-represented accents, consistent with the disparities in Table~\ref{tab:table1}. ERM reduces error rates but leaves several accents with high WER. Our proposed fusion strategy improves performance across all accents for Whisper, and for Seamless, it improves almost all accents, with the exception of Urdu and Indian English. This limitation may stem from the under-representation of these accent groups in the initial pre-training, and we plan to investigate this further in future work.

\vspace{-0.4cm}
\subsection{Analysis}

\begin{figure}[t]
\centering
\includegraphics[height= 140pt,width=210pt]{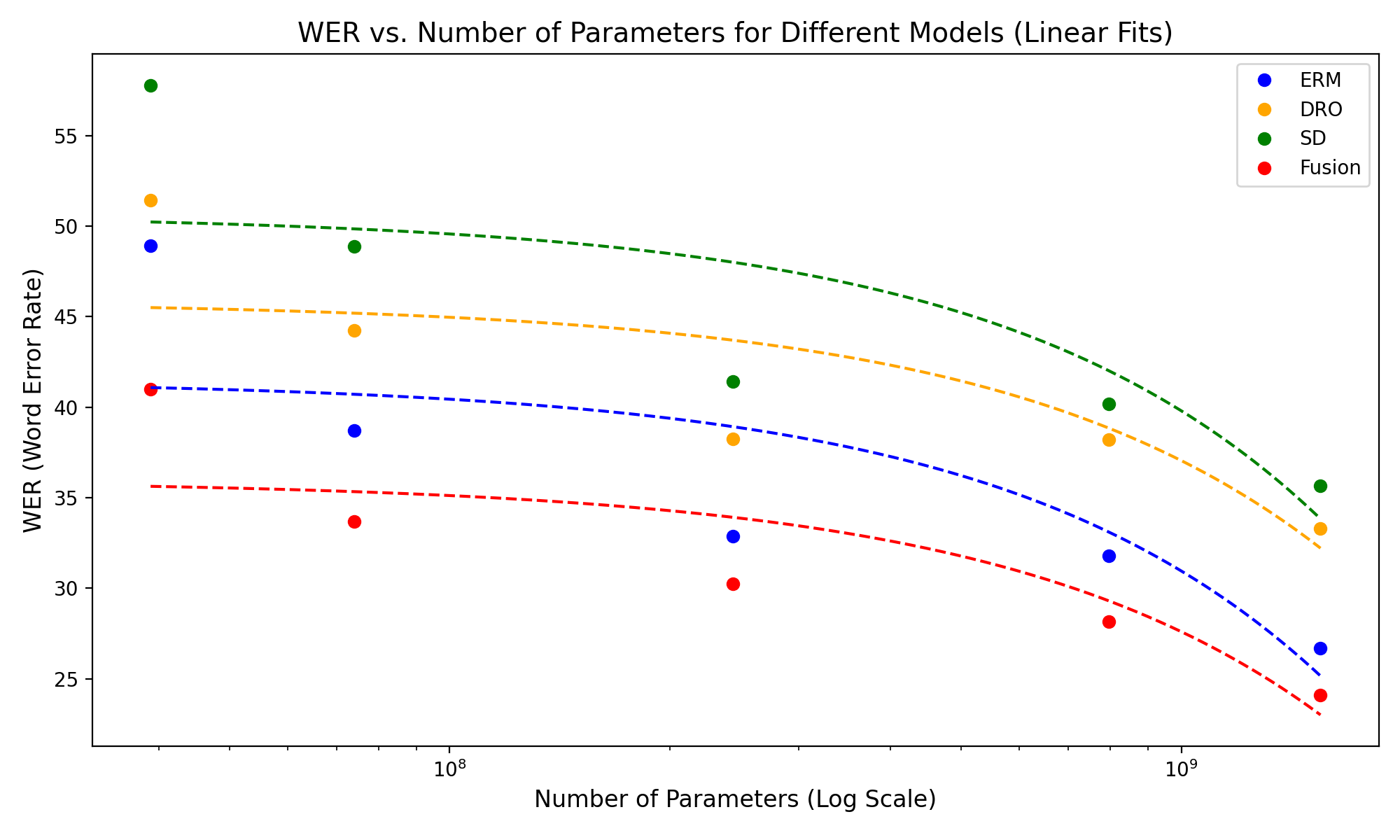}
\vspace{-6pt}
  \caption{WER across Whisper models sizes.}\label{fig:5}
  \vspace{-0.5cm}
\end{figure}

\textbf{Effect of Model Size:} 
The impact of fairness-promoting algorithms on performance decreases as models grow larger. Larger models reduce WER across all strategies, but the gap between ERM, DRO, SD, and Fusion narrows with scale. This observation, shown in Figure~\ref{fig:5}, suggests that scaling itself mitigates some fairness trade-offs while raising new research for sustainable language modeling.

\textbf{Typological Distance:} 
We consistently observed the highest performance in Asian accents, followed by European accents (e.g., Romanian), and African accents, with some more European varieties (e.g., Italian). We found no clear correlation with typological distance. For example, Romanian and Italian are closely related but differed by nearly 20 WER points.

\textbf{Word Length:}
We also computed the average word length in the test data for each variety of English to control for this potential confounding factor. However, we found no correlation between word length and WER.


\section{Conclusion}

We investigated the performance gaps of widely adopted ASR models such as Whisper and Seamless on English accented by L2 speakers. To address these disparities, we applied fairness-promoting objectives—SD, DRO, and IRM—for fine-tuning. Our results show that fine-tuning with fairness objectives consistently improves macro-averaged WER compared to both pretrained models and standard empirical risk minimization. Moreover, our proposed fusion of ERM with fairness objectives further enhances performance, outperforming individual objectives. We also observed that increasing model size reduces macro-averaged WER, highlighting the benefit of scaling. Although we explored potential correlations between WER and factors such as native language typological distance and average word length, we did not find strong evidence of direct relationships. Nevertheless, our fine-tuning approach delivered consistent improvements across most L2 accents, with only a few challenging cases such as Urdu and Indian English in the case of Seamless. In future work, we will focus on these difficult accents to further reduce the performance gap and advance towards more equitable ASR systems.


\label{sec:refs}
\small

\bibliographystyle{IEEEtran}
\bibliography{strings,refs}

\end{document}